# Fast Retinomorphic Event-Driven Representations for Video Recognition and Reinforcement Learning

**Wanjia Liu**
Rice University

**Huaijin Chen**
Rice University

**Rishab Goel**
IIT-Delhi

**Yuzhong Huang**
Olin College of Engineering

**Ashok Veeraraghavan**
Rice University

**Ankit Patel**
Baylor College of Medicine & Rice University

## Abstract

Good temporal representations are crucial for video understanding, and the state-of-the-art video recognition framework is based on two-stream networks [61, 17]. In such framework, besides the regular ConvNets responsible for RGB frame inputs, a second network is introduced to handle the temporal representation, usually the optical flow (OF). However, OF or other task-oriented flow [68] is computationally costly, and is thus typically pre-computed. Critically, this prevents the two-stream approach from being applied to reinforcement learning (RL) applications such as video game playing, where the next state depends on current state and action choices.

Inspired by the early vision systems of mammals and insects, we propose a fast event-driven representation (EDR) that models several major properties of early retinal circuits: (1) logarithmic input response, (2) multi-timescale temporal smoothing to filter noise, and (3) bipolar (ON/OFF) pathways for primitive event detection [12]. Trading off the directional information for fast speed (> 9000 fps), EDR enables fast real-time inference/learning in video applications that require interaction between an agent and the world such as game-playing, virtual robotics, and domain adaptation. In this vein, we use EDR to demonstrate performance improvements over state-of-the-art reinforcement learning algorithms for Atari games, something that has not been possible with pre-computed OF. Moreover, with UCF-101 video action recognition experiments, we show that EDR performs near state-of-the-art in accuracy while achieving a 1,500x speedup in input representation processing, as compared to optical flow.

## 1 Introduction

Deep learning and related techniques have resulted in substantial advances in image understanding over the last decade [36, 34, 28], resulting in a new-found sense of optimism regarding possibilities in many application areas, including autonomous robots and self-driving cars. Unfortunately, the current state of practice in video understanding tasks is either to (a) process each frame in the video sequence independently, ignoring the temporal structure, or (b) pre-compute the temporal representations, as seen in the two-stream architecture for video recognition [61]. The latter is efficient but necessarily precludes applications like reinforcement learning, where the next state depends on the current state and the actions taken by the agent. Therefore, if the two-stream approach were to be used in RL, the temporal representation will have to be computed on-the-fly. Take the Atari Centipede gameplay as an example, reinforcement learning usually converges to the optimal strategy after 70 millions steps of trials and errors. At each steps, assuming 4 frames of video frames are considered to decide an action, then under the conventional OF-based two-stream framework, there will be $70 \times (4-1) = 210M\ frames$ of OF field to be calculated across the entire training process on-the-fly. Even if we use a relatively fast OF algorithm (i.e. TV-L1 [51] which runs at $30 fps$ for $320 \times 240$ inputs), it will still take more than $210M/30 = 7M\ secs = 81\ days$ to just compute the OF! As

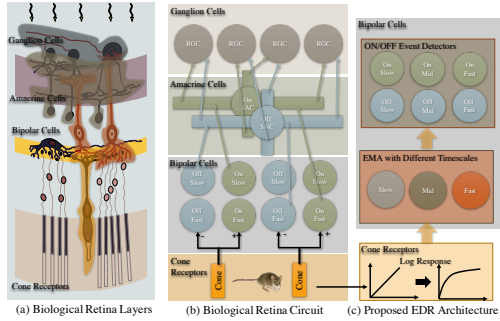
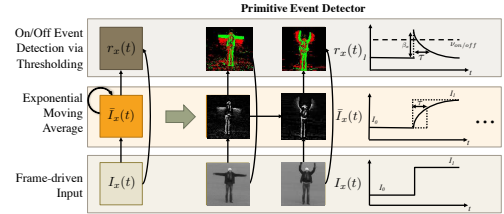

(a) **Neural circuit of biological early motion vision system vs. EDR:** We focus on three key features of the retina: (1) Logarithmic response, (2) ON/OFF event pathway, and (3) integration of multiple timescale events. The left-most column is the biological neural circuit, the middle column (adapted from [12]) is the abstracted circuit, and the right most column is our EDR implementation of the abstracted circuit.

(b) **EDR as a Retinomorphic Primitive Event Detector:** From the frame-driven inputs, we first apply an exponential moving average filter to smoothen the event estimation. We then calculate the relative changes/return in the input stream. Finally we threshold the relative changes/return and generate binary On/Off events.

Figure 1: Overview of the Proposed EDR

such, there are still significant gains to be captured in speed, accuracy, bandwidth and energy, by explicitly leveraging the temporal redundancy structure in video.

Why haven't we seen such progress? We believe there are two principal reasons for the slow pace of progress in video understanding using deep networks. First, the massive size and data rates needed in video, make even the simplest feed-forward processing computationally challenging to accomplish, especially in settings like game-playing where real-time processing is needed. A natural solution to this problem is to use event-driven sensor or input representations, which bring orders of magnitude reduction in sensor power consumption and data bandwidth, which are both especially helpful on energy-constrained platforms. However, existing efforts in this vein focus mostly on optimizing network architecture [30]. Second, a good input representation for temporal dynamics – one that accounts for and exploits the inherent redundancy – is crucial to good performance in video understanding. For example, optical flow has significant added value in video understanding tasks, but it is expensive to compute, making it impossible for many tasks that requires real-time video input, such as reinforcement learning (Figure 2a and 2b). As a result, despite recent breakthroughs in static image understanding [28, 34], much more research is needed in developing new fast, temporally-aware representations for video understanding.

**Contributions** In this paper, we propose a retinomorphic learnable event driven representation (EDR) for video. Our EDR has tunable parameters, enabling us to capture meaningful task-relevant events from the data. In the proposed EDR, we captured some of the simple properties of the

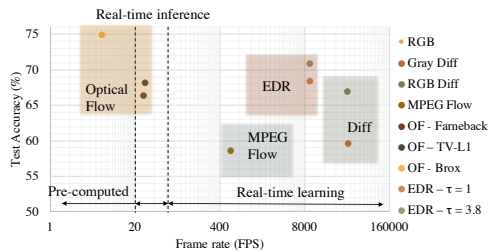

(a) **Speed Accuracy Trade-off for Temporal Representations:** We compare the proposed EDR with serveral common input representation for temporal dynamics. The major features of EDR are 1) larger temporal receptive field (RF) for capturing events of different timescales, 2) temporal smoothing to handle the noise, 3) soft thresholding that allows distinguish events while keep some of the texture of moving objects, and 4) Potentially learnable parameters.

| | Optical Flow | Inter-frame Differences | Event-driven Camera (DVS128) | EDR |
|---|---|---|---|---|
| Feature | Patch Displacement | Pixel intensity change | Pixel intensity change | Pixel intensity change |
| Computation | High | Low | Low | Low |
| Run-time | Pre-computed | Real-time | Real-time | Real-time |
| Temporal RF | 2 frames | 2 frames | Multiple Timesteps | $\tau$ frames |
| Temp. Smooth | No | No | No | Yes |
| Spatial RF | >10 pix | 1 pix | 1 pix | 1 pix |
| Spatial Smooth | Yes | No | No | No |
| Thresholding | No | Hard | Hard | Soft |
| ON/OFF Channels | No | No | Yes | Yes |
| Learnable Parameters | No | No | No | Yes |

(b) Quantitative and Qualitative Comparison of Different Temporal Representations

Figure 2: EDR's Major Features Compared to Other Temporal Representations

biological retina and the visual cortex. Our implementation is preliminary and we only conduct first-order studies of EDR, and yet the performance gains (near state of art classification performance with potentially order of magnitude reductions in data throughput and power requirements) strongly suggest that retinomorphic event-driven representations are worthy of further study. The main contributions of our work are:

1. We propose a retinomorphic event-driven representation (EDR) for video, instantiated as an RNN layer, that realizes three important functions of the biological retina: logarithmic transformation, ON/OFF pathways for event detection and the integration of multiple timescales.
2. We show systematic improvements in the performance (accuracy, response time, latency, throughput) and learning speed for our proposed learned EDRs as compared to FDRs on the task of Atari game playing via reinforcement learning and UCF-101 action recognition.
3. We are able to apply the EDR model to an event-driven camera hardware, showcasing a joint software/hardware event driven motion vision system that performs high-level visual understanding tasks with order of magnitude reductions in data throughput and power per voxel compared to traditional-image-sensor based implementations. (See Appendix)

## 2 Related Work

**Biological Retinas.** In order to survive in a hostile environments, animals have evolved visual systems with fast reaction times, and the retina is the first stage of processing. The retina has many interesting properties both in its structure and in its variety of responses. However, there is reasonable consensus [12, 54] that at least these five major features are essential for motion-based vision (see Figure 1a): (1) Photoreceptors have a logarithmic response to luminance input; (2) event detection happens in parallel ON/OFF pathways in multiple cell types (e.g. bipolar and ganglion cells) where ON/OFF events are generated based on spatiotemporal changes in luminance [23, 64, 48]; (3) Fast/slow pathways for distinguishing/integrating events across different time scales [4]. (4) Primitive motion detection for the cardinal directions (e.g. LPTC or DSGC cell types [44, 69]); and (5) 4-channel color vision cone (RGB) and rod (grayscale) cells [9]. Several recent studies propose computational neuroscience models for the retina and learning process [27, 42], but those models were not designed for deep practical learning tasks. This is by no means an exhaustive list; indeed the structure and function of the retina are active areas of ongoing research. In this paper, we focus on the event-driven nature of the retina, instantiating and exploring properties (1), (2) and (3) above, and leaving others – listed or not – for future work.

**Retinomorphic Cameras.** Research on neuromorphic image sensor and systems can be trace back to decades ago [10, 2, 22]. Along the way, we have seen impressive research prototypes [11, 53], including a sensor that reproduces all five layers of the retina [71]. Recently, retinomorphic event driven image sensors have been commercialized, such as the Dynamic Vision Sensor (DVS) [38], Dynamic and Active-pixel Vision Sensor (DAVIS) [13, 8] and Asynchronous Time-based Image Sensor [53]. Unlike conventional cameras that output grayscale or color intensities for each pixel and each frame, event cameras detect and report only significant changes in intensity at each pixel. As a result, DVS is capable of dramatically higher throughput and lower latency as compared to a conventional camera. We have seen recent applications of event-driven camera in many areas of computer vision and robotics: such as structured light active 3D imaging [43], multi-view stereo [55], high-speed tracking [47], panoramic tracking [57], face detection and intensity reconstruction [7], visual odometry [16, 56], motion flow [6] and real-time 3D scene reconstruction [33].

Despite the amount of work on using event cameras for low-level vision tasks, there still remains a gap between event cameras (e.g. artificial retinas) and high-level semantic understanding (e.g. artificial cortex). An early exploration was carried out by [52], in which the authors attempted to map DVS camera output events back to frames, and then feeding the frames to a 5-layer ConvNet to tell which one of the four directions the object is moving towards.

**Neural Network for Temporal Sequences** Recent works in video recognition using both deep learning [3, 70, 61] and traditional hand-designed features [32, 58, 21], suggesting that efficiently modeling/integrating information across time is very important for performance. Recurrent Neural Networks (RNNs) with Long- and Short-term Memory (LSTM) [26, 29] – employed successfully for speech recognition and natural language processing – might be a promising model for such problems. Marrying convnets to LSTMs resulted in the long-term recurrent convnet (LRCN), which has shown some promising but not outstanding results in video recognition[20]. LRCNs begin to integrate information across time but do so at later stages in the visual processing architecture. This is in stark contrast to the biological visual system, wherein temporal feature detection and event generation happen at the *earliest* stages yielding an architecture that is *event-driven from end to end*.

Sigma-delta quantized networks (SDQN)[49] also investigate sparse temporal representations. However, the main goal of SDQNs is reducing computation (FLOPs) by passing quantized activation differences through time. EDR, on the other hand, is an event-driven input representation, looking for meaningful task-specific changes in the input.

**Two-Stream Architectures for Video Inference** Recently, deep architectures have been used to obtain better performance on standard video inference benchmarks, such as UCF-101[63] and HMBD-51[35]. The state-of-the-art methods for those benchmarks are based on the two-stream ConvNets framework proposed by Simonyan, et al. [61]. Such framework uses two separate ConvNets to handle RGB frames and optical flow derived from the RGB frames, respectively. The final prediction is based on the consensus of both networks. The success of the two-stream framework demonstrated the importance of a good temporal representation, however, the temporal representation itself is much less discussed in many two-stream-based frameworks. Even the state-of-the-art two-stream frameworks, such as the Temporal Segmentation Networks (TSN) [67], Long-Term ConvNets(LTC) [65] and Inflated 3D ConvNets (I3D) [17], simply use off-the-shelf algorithms to precompute a costly but accurate optical flow (OF) field. Such accuracy is costly to compute, but may not be necessary, as suggested by recent work [68, 24, 72]. Instead, a task-driven flow can be jointly learned end-to-end as a fast alternative to conventional OF. Nevertheless, the task-driven flow can only be computed at frame rates of up to 12-120 fps, which is still prohibitively slow for training, especially for applications like RL in video games.

**Reinforcement Learning-based Video Game Play** Deep networks have been very successful in solving reinforcement learning video game problems and demonstrated great capability in environment transition and agent behavior modeling. To name a few, Minh et al. [46] first introduced Deep Q-learning (DQN) by approximating Q function using a neural network and enabled Q-learning to achieve near or over human performance on Atari video games; Minh et al. [45] approximates advantage and policy vector using single network and have multiple agents asynchronously explore environment, largely improved performance over DQN. Build on top of A3C, Parallel Advantage-Actor-Critic (PAAC) [18] implements A3C on CPU and shortens A3C training time to one day. However, recent RL improvements are mostly on RL strategy, problem formulation and network structure. To our knowledge, none of existing work are able to utilize the two-stream framework for RL due to the slow OF computation. We believe we are the first to investigate the event-driven temporal input representation for RL, where all the existing RL approaches are largely relying on RGB or grayscale frame-based input.

## 3 Event-Driven Representation

We propose a simple event detector: a *thresholded exponential moving average (tEMA)* of (relative) changes in the input. Despite its simplicity, this simple detector is widely used as a temporal event detector and descriptor in many areas including high-frequency finance [40] and the mammalian retina [62, 50]. In both cases, fast response times are critical. We now describe the structure of the tEMA which is composed of three components: (1) an exponential moving average filter, (2) a relative change computation and (3) a thresholding operation. We show the conceptual diagram of such procedure in Figure 1b.

**Exponential Moving Average (EMA).** For each pixel location $x$, the pixel intensity $I_x(t)$ is noisy and variable. In order to smoothen the estimate, we apply an exponential moving average (EMA) filter to the sequence $I_x(t)$ to get the filtered sequence $\bar{I}_x(t; \tau_{1/2}) \equiv EMA(I_x(t); \tau_{1/2})$ where the half-life parameter $\tau_{1/2} \in \mathbb{R}_+$ controls the memory of the filter. Intuitively, a new data point affects the EMA for $\tau_{1/2}$ timesteps before decaying into half of starting amplitude. Effectively, an EMA weighs the recent past is *exponentially more* than the distant past.

One computational advantage of the EMA is that it can be computed recursively as

$$\begin{aligned}\bar{I}_x(t) &= \bar{I}_x(t-1) + \alpha(I_x(t) - \bar{I}_x(t-1)) \\ &= (1-\alpha)\bar{I}_x(t-1) + \alpha I_x(t),\end{aligned} \quad (1)$$

where $\bar{I}_x(t)$ is the EMA of input $I_x(t)$ at time $t$ and pixel location $x$ and we have suppressed the dependence on the memory parameter $\tau_{1/2}$. In fact, the dependence on $\tau_{1/2}$ will be indirectly specified through another tunable parameter $\alpha \equiv 1 - 2^{-1/\tau_{1/2}} \in [0, 1]$. A larger $\alpha$ (smaller $\tau_{1/2}$) places more weight on the most recent inputs and thus forgets earlier inputs $I_x(t)$ more quickly. Note that this recursive update for the EMA is linear in $I_x(t), \bar{I}_x(t)$ and so can be implemented as a *linear recurrent neural network* (RNN).

**Relative Changes/Returns.** We next need to define a way to compute changes in the input stream. One simple approach is to compute the relative change of the input with respect to past inputs i.e. a return. Given a smoothed estimate $\bar{I}_x(t)$ of the input stream, this return stream is defined as

$$R_x(t) \equiv \left(\frac{I_x(t)}{\bar{I}_x(t)}\right)^{\beta_x} \qquad \text{or} \qquad r_x(t) \equiv \beta_x \ln\left(\frac{I_x(t)}{\bar{I}_x(t)}\right).$$

Note that returns are dimensionless measures of changes in the input stream just like e.g. stock price returns in finance. Intuitively, when the input stream is constant $I_x(t) = I_0$ the return stream $r_x(t) \to 1$ since the EMA $\bar{I}_x(t) \to I_0$ in $O(\tau_{1/2})$ timesteps. If the input stream is an impulse, the return stream jumps quickly to its peak response and then decays with half-life $\tau_{1/2}$ back to 1 (see Fig. 1b). The amplitude of the peak response is controlled by $\beta_x \in \mathbb{R}$. Intuitively, increasing/decreasing $\beta_x$ makes the event detector more/less sensitive to changes in the input (e.g. pixel intensity changes in the scene).

**Event Detection via Thresholding.** Given a sequence of real-valued returns, we now define a simple event detector via a thresholding operation that determines whether a change is "significant", analogous to a noise floor in a signal detection problem.

Our *soft thresholding* operation employs a *bipolar* structure, inspired by the retina, that detects two kinds of input events: ON and OFF. The output event streams are mathematically defined as

$$E_{x,ON}(t) \equiv [r_x(t) - (1 + \nu_{ON})]_+ \in \mathbb{R}_+ \tag{2}$$

$$E_{x,OFF}(t) \equiv [r_x(t) - (1 - \nu_{OFF})]_- \in \mathbb{R}_+, \tag{3}$$

where $[b]$ is defined as 1 if statement $b$ is true and 0 if it is false, and $[r]_+ \equiv ReLU(r) \in \mathbb{R}_+$, $[r]_- \equiv ReLU(-r) \in \mathbb{R}_+$ are the positive and negative parts, respectively, of the real number $r \in \mathbb{R}$. The threshold parameters $\nu_{ON}, \nu_{OFF}$ determine how large a relative change in the inputs is required for an ON/OFF event to be generated. For example, if $\nu_{ON} = +5\%$ then a 5% increase in the input $I_x(t)$ relative to its EMA $\bar{I}_x(t)$ is needed in order for an ON event to be generated. A similar relationship holds for $\nu_{OFF}$. This mimics the retinal firing rates after the ON/OFF event is detected.

Note the similarities and differences between the bipolar events and a standard weighted ReLu layer in a recurrent ConvNet. Both can be written as recurrent weight layers with a biased ReLu. Despite this similarity, there are a few key differences inspired directly from the retina. First, our EDR possesses bipolar (ON/OFF) semantics, i.e. there are two parallel channels whose purpose is to detect significant changes in the two possible directions. Second, the weight and bias $\beta_x, \nu_{x,s}$ are *interpretable* as sensitivity parameters and detection thresholds, respectively. Third, The input into the bipolar ReLu is a hand-designed feature – the log-returns stream of the inputs – that is hand-designed to be a trend detector. The log – inspired by retinal photoreceptor responses – enables the processing of inputs with large dynamic range. (This primitive event detector is also commonly used in high-frequency finance.)

**Multiple Timescale Events** Biological retina has synapses/connections that combines events from fast and slow pathways to form a event stream that is sensitive to different time scales. We mimic this by providing log-return from short, medium, and long time scales $\{\alpha_S, \alpha_M, \alpha_L\}$ EDR. Note that the different timescales have different (learnable) weights $\{\beta_S, \beta_M, \beta_L\}$ associated.

$$\boldsymbol{r_x}(t) \equiv \{r_x(t; \alpha_j)\}, j \in \{S, M, L\} \tag{4}$$

**Temporal dynamics representation comparison** There are several input representations that aim for capture the temporal dynamics, namely optical flow[15, 25], DVS[38] camera hardware, simple inter-frame difference. We compare the EDR with them, and summarize the differences in Table 2b. The major features of EDR are 1) larger temporal receptive field (RF) for capturing events of different timescales, temporal smoothing to handle the noise and soft thresholding that allows distinguish events while keep some of the texture of moving objects. We also plot out the response of different input representation across time for a random pixel location in the first "Archery" clip of the UCF-101 datasets in Figure 3b. The visualization of corresponding frames can be found in Figure 3a.

## 4 Experiments and Results

To evaluate and compare the performance of our proposed retinally-inspired input representations in a variety of application scenarios, we carry out experiments on Atari gameplay reinforcement learning (Section 4.1), as well as UCF-101 (Section 4.2) action recognition. Due to the limited space, implementation details are discussed in Appendix C and D. In addition, we show hardware experiments in Appendix A, and used a smaller action recognition datasets KTH to perform ablation studies on our proposed EDR framework (Appendix B).

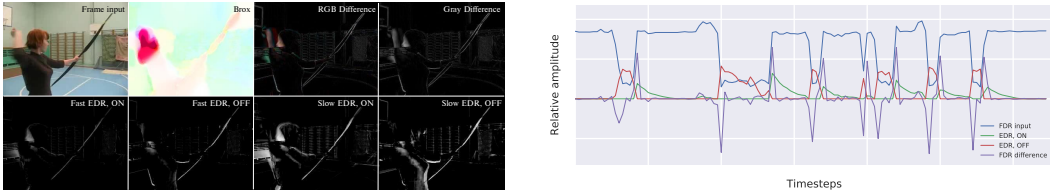

(a) Visualization of different input modalities of a UCF-101 frame sequence.

(b) Visualization of different input modalities for a random pixel location of a UCF-101 frame sequence.

Figure 3: **Temporal Input Representations Comparisons on UCF-101 Frames:** 3a. We visualize a sample frame from UCF101 archery class. First row from left to right: RGB, Brox flow, RGBDiff, GrayDiff. Second row from left to right: Fast decay EDR ON, Fast decay EDR OFF, Slow decay EDR ON, Slow decay EDR OFF. Notice the stationary lattice pattern in the background is less emphasized in fast decay EDR than Difference, while being responsive to the archer. 3b. Output value of different representation across time for a randomly chosen pixel location in the first archery clip in the UCF-101 datasets are shown.

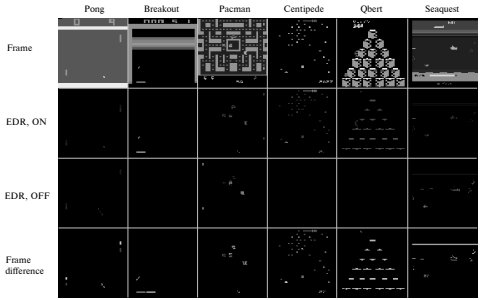

Figure 4: **Temporal Input Representations Comparisons on RL Atari Game Play:** Original frame, EDR and Diff representations for 6 tested Atari games are shown.

### 4.1 Atari Game Reinforcement Learning (RL)

**RL Environment and Algorithm** A natural choice for testing the proposed EDR is RL-based game play tasks, because games involve rich visual action interactions, and visual-event-action is a common reflex found in animals that take use of event driven representation. We use a random subset of six (out of 55) Atari 2600 console games in Atari Learning Environment (ALE) [14], namely Pong, Breakout, Pacman, Centipede, Qbertand and Seaquest, ranked roughly based on game difficulty from easy to hard. Our baseline RL setup is Parallel Advantage-Actor-Critic (PAAC) [18]. PAAC is a synchronous implementation of state-of-the art algorithm A3C [45] in GPU. In the baseline setup, game playing frames are feed into controller network directly. We concatenate EDR to the original FDR input for all our EDR comparison experiments, where the details are described in Appendix C.

**Network Architecture and Training Procedure** Controller network architecture [18] is important for both extracting features and mapping them into value and policies, and affects agent performance. We refer the reader to Appendix C for the details about the network architecture and RL policies.

|           | FDR    |       |       | FDR+EDR |       |       |
|-----------|--------|-------|-------|---------|-------|-------|
|           | Avg.   | Best  | Worst | Avg.    | Best  | Worst |
| Pong      | 18.3   | 21    | 14    | **19.5**| 21    | 14    |
| Breakout  | 422.5  | 494   | 378   | **446.6**| 842  | 365   |
| Pacman    | 2828.6 | 6682  | 456   | **3719.7**| 5030 | 1850 |
| Centipede | 1523.2 | 4243  | 302   | **2828.6**| 6682 | 456  |
| Qbert     | 17088.8| 19225 | 11775 | **19560**| 22800| 19025 |
| Seaquest  | **1691**| 1700 | 1280  | 1674    | 1760  | 1360  |
| Avg. gain | -      | -     | -     | **+23.8%**| +20.9% | +70.2% |

Table 1: **Atari RL Experiment Results:** The min, max and mean total reward score of 7 different games over 30 different plays are shown. Comparison are made between the conventional FDR inputs and proposed EDR + FDR inputs. Notice that worst case performance for EDR + FDR is much better than that of FDR alone.

**RL Experiment Results** Experimental results of 6 different games are shown in Table 1, where the min, max and mean total reward score of over 30 different test runs are reported. We compare

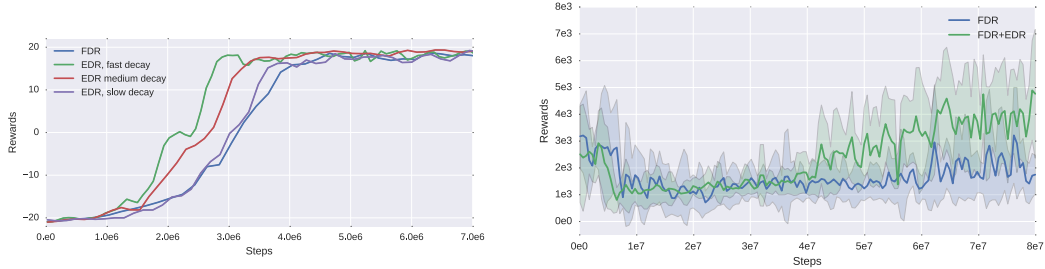

(a) 'Pong' Training Curve for Different EDR Time Scales

(b) 'Centipede' Training Curve

Figure 5: **RL Training:** 5a. Rewards over game steps at training are shown for Pong. Higher is better. EDR+FDR learns 2x faster than the FDR alone. The strategy learned by the FDR+EDR is superior to that learned by FDR, requiring only 1 shot to defeat the opponent as compared to multiple shots for FDR. 5b. Rewards over game steps at training are shown for Centipede. EDR+FDR reaches 1.8x higher reward than FDR alone.

the total reward scores of FDR + EDR inputs with those of conventional FDR inputs. It is clear that added EDR inputs greatly improve the performance. Meanwhile, we visualize the EDR stream in Figure 4, where we can see that EDR picks up short-term events that are relevant to wining the game.

Plotting a sample training process for Pong game in Figure 5a, we found that both FDR and FDR+EDR networks achieve the maximum score of 21, but EDR+FDR can learn 1.7x faster (3M vs. 5M training episodes to reach a high score of 18) than the FDR alone. Furthermore, the strategy learned by the FDR+EDR is superior to that learned by FDR, requiring only 1 shot to defeat the opponent as compared to multiple shots for FDR. Figure 5b is an example training process for Centipede game. We found that the addition of the EDR representation accelerates training in PAAC. Moreover, we find that FDR+EDR based training reaches 1.8x higher reward than FDR alone and is still increasing when training terminates. FDR+EDR can reach a maximum reward of 8,300 later on in training while FDR struggles to reach 3,500. More training curves of different games can be found in the slides in supplementary material.

### 4.2 UCF-101 Action Recognition Experiments

Here we compare EDR with temporal representation alternatives on Long Term Convolutional [65] (LTC) architecture. We would like to emphasis that the goal of this experiment is not to compete with the state-of-the-art methods in action recognition, but rather, understand how EDR with different parameters perform compared to other existing temporal input representations on state-of-the-art baseline network (i.e. LTC). We use the following representations in comparision: RGB, MPEG Flow, difference between grayscale intra-frame differences (GrayDiff), RGB intra-frame differences (RGBDiff), TV-$L^1$ OF [51], Farneback OF [25], Brox OF [15], fast decay soft threshold EDR ($\alpha = 0.5$) and slow decay soft threshold EDR ($\alpha = 0.166$).

**Datasets** UCF-101 consists of 101 action categories of more than 13k video clips in 30fps. Each clip is in 320x240 resolution and the average mean clip length is 7.21 seconds. UCF101 contains videos with rich camera motion and cluttered background, and comes with 3 pre-determined train/test splits. All splits are designed to balance out video nuisances such actors and background, ensuring fairness between training and testing sets. We also perform standard data augmentation when loading the data, where the details can be found in Appendix D. We report performance on the first splits in all our experiments.

**Network Architecture** LTC architecture has five 3D convolutional layers with 3x3x3 filter size, ReLu nonlinearity and volumnetric max pooling, followed by 2 fully connect layers with 2048 units and dropout. LTC originally compares different input modalities using 60 frame architecture. LTC reaches highest 92.3% accuracy on UCF101 by pre-training on Sports-1M, and combining Brox OF and Improved Dense Trajectories (IDT) along with the RGB input. Here we use same setup to enable fair comparison and save computation cost. For the details about the LTC Network, we refer the reader to the original paper [65] and Figure 12 in Appendix D for the end-to-end network diagram.

**UCF101 Experiment Results** A summary of training results on the baseline LTC network is shown in Table 2. When training from scratch on UCF-101 data, among all the temporal representations, the best performing temporal representation is the Brox OF, which achieve 74.8% accuracy on UCF-101. Overall, EDR (70.7%) performs better than FDR, MPEG Flow[32], Farneback OF [25], TV-$L^1$ OF [51], and with a much faster computation speed measured by frame-per-second compared with all optical flow variants. Computing intra-frames difference (RGBDiff and GrayDiff) is indeed 4x faster than EDR, but the accuracy is notcieably lower. Nevertheless, 9.8K fps on EDR is fast enough to enable real-time applications that are not practical for optical flows. We can safely conclude that EDR can trade-off accuracy slightly (and is still among the best performing temporal representations) for much better computing efficiency.

| Type | Representations | Accuracy | Speed (FPS) |
| --- | --- | --- | --- |
| Frame | RGB | 57.0 [65] | - |
| Frame Diffs | Gray | 59.5 | 37.6K |
| | RGB | 66.8 | 36.9K |
| Flow | MPEG Flow [32] | 58.5 [65] | 591.8 [32] |
| | OF- Farneback [25] | 66.3 [65] | 27.3 |
| | OF - TV-$L^1$ [51] | 68.0 | 29.1 [51] |
| | OF - Brox [15] | **74.8** [65] | 6.3 |
| EDR | $\tau_{1/2} = 1$ frame | 68.3 | 9.8K |
| | $\tau_{1/2} = 3.8$ frames | **70.7** | 9.8K |

Table 2: **Comparing Different Temporal Representation on the UCF-101 Action Recognition Tasks:** We show the test accuracy for different input representations in the UCF-101 action recognition test. We found that EDR perform better than simple inter-frame difference and more computationally expensive Farnbecks optical flow. However, worse than more accurate optical flow method proposed by Brox.

Figure 6 compares number of samples correctly classified from EDR pre-processed network and Brox pre-processed network, sorted by the difference between the two, from large to small. Overall, EDR and Brox performs similarly: both found PizzaTossing as a difficult class and BenchPress as a simple one. However in a small portion of classes the diffences are significant. Example visualizations of the EDR response on UCF-101 video sequence can be found in Figure 13 in Appendix D.

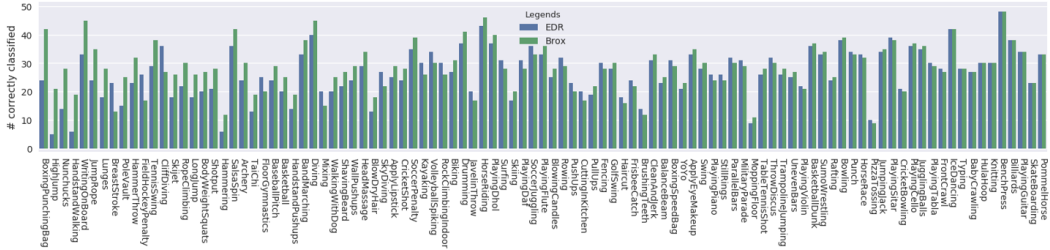

Figure 6: **UCF-101 Results - EDR vs. RGB FDR:** Number of correctly classified videos per class between EDR and Brox[15]. Blue is EDR, green is Brox, sorted by the difference between the two for each class.

## 5 Conclusions

In summary, we propose EDR, an event-driven retinomorphic input representation, for extracting temporal dynamics in video. We show that EDR improves performance in Atari game playing reinforcement learning and dramatically improves the speed-accuracy tradeoffs in UCF-101 action recognition tasks, compared with alternative temporal representations. Furthermore, as shown in Appendix A, we build an early-stage prototype consisting of DVS event camera hardware and a light-weight feature-tracking RCNs for performing standard action recognition tasks with orders of magnitudes reduction in sensing power and data transmission rates. Meanwhile, we acknowledge that our current EDR architecture is an early proof-of-concept and has some limitations. We have not yet implemented some important properties of the retina. For instance, (1) We do not have a stable learnable EDR architecture. In our preliminary experiments, end-to-end learning the EDR parameters seems to be quite unstable, probably due to the fact that EDR is formulated as an RNN, and the Back-propagation-thru-Time (BPTT) algorithm used to train it yields unstable gradients. (2) We do not include transient and sustained cell types, resulting in some loss of adaptability that is inherent in the retina. (3) We do not simulate any spatial processing in the retina (e.g. Horizontal and Amacrine cells), therefore our events are based solely on temporal changes, rather than spatiotemporal changes. We hope our preliminary studies on EDR will motivate more research in this direction.

# References

[1] Haiam A Abdul-Azim and Elsayed E Hemayed. Human action recognition using trajectory-based representation. *Egyptian Informatics Journal*, 16(2):187–198, 2015. 15

[2] Andreas G Andreou, Richard C Meitzler, Kim Strohbehn, and KA Boahen. Analog vlsi neuromorphic image acquisition and pre-processing systems. *Neural Networks*, 8(7):1323–1347, 1995. 3

[3] Moez Baccouche, Franck Mamalet, Christian Wolf, Christophe Garcia, and Atilla Baskurt. Sequential deep learning for human action recognition. In *International Workshop on Human Behavior Understanding*, pages 29–39. Springer, 2011. 3

[4] Tom Baden, Philipp Berens, Matthias Bethge, and Thomas Euler. Spikes in mammalian bipolar cells support temporal layering of the inner retina. *Current Biology*, 23(1):48–52, 2013. 3

[5] Nicolas Ballas, Li Yao, Chris Pal, and Aaron Courville. Delving deeper into convolutional networks for learning video representations. *arXiv preprint arXiv:1511.06432*, 2015. 13, 14

[6] Patrick Bardow, Andrew J. Davison, and Stefan Leutenegger. Simultaneous optical flow and intensity estimation from an event camera. In *The IEEE Conference on Computer Vision and Pattern Recognition (CVPR)*, June 2016. 3

[7] Souptik Barua, Yoshitaka Miyatani, and Ashok Veeraraghavan. Direct face detection and video reconstruction from event cameras. In *Applications of Computer Vision (WACV), 2016 IEEE Winter Conference on*, pages 1–9. IEEE, 2016. 3, 13

[8] Raphael Berner, Christian Brandli, Minhao Yang, Shih-Chii Liu, and Tobi Delbruck. A 240× 180 10mw 12us latency sparse-output vision sensor for mobile applications. In *VLSI Circuits (VLSIC), 2013 Symposium on*, pages C186–C187. IEEE, 2013. 3

[9] HR Blackwell and OM Blackwell. Rod and cone receptor mechanisms in typical and atypical congenital achromatopsia. *Vision Research*, 1(1):62–107, 1961. 3

[10] Kwabena Boahen. Retinomorphic vision systems. In *Microelectronics for Neural Networks, 1996., Proceedings of Fifth International Conference on*, pages 2–14. IEEE, 1996. 3

[11] Kwabena Boahen. Neuromorphic microchips. *Scientific American*, 292(5):56–63, 2005. 3

[12] Alexander Borst and Moritz Helmstaedter. Common circuit design in fly and mammalian motion vision. *Nature neuroscience*, 18(8):1067, 2015. 1, 2, 3

[13] Christian Brandli, Raphael Berner, Minhao Yang, Shih-Chii Liu, and Tobi Delbruck. A 240× 180 130 db 3 $\mu$s latency global shutter spatiotemporal vision sensor. *IEEE Journal of Solid-State Circuits*, 49(10):2333–2341, 2014. 3

[14] Greg Brockman, Vicki Cheung, Ludwig Pettersson, Jonas Schneider, John Schulman, Jie Tang, and Wojciech Zaremba. Openai gym, 2016. 6

[15] Thomas Brox, Andrés Bruhn, Nils Papenberg, and Joachim Weickert. High accuracy optical flow estimation based on a theory for warping. *Computer Vision-ECCV 2004*, pages 25–36, 2004. 5, 7, 8, 18

[16] Cesar Cadena, Luca Carlone, Henry Carrillo, Yasir Latif, Davide Scaramuzza, José Neira, Ian Reid, and John J Leonard. Past, present, and future of simultaneous localization and mapping: Toward the robust-perception age. *IEEE Transactions on Robotics*, 32(6):1309–1332, 2016. 3

[17] Joao Carreira and Andrew Zisserman. Quo vadis, action recognition? a new model and the kinetics dataset. In *2017 IEEE Conference on Computer Vision and Pattern Recognition (CVPR)*, pages 4724–4733. IEEE, 2017. 1, 4

[18] Alfredo V Clemente, Humberto N Castejón, and Arjun Chandra. Efficient parallel methods for deep reinforcement learning. *arXiv preprint arXiv:1705.04862*, 2017. 4, 6, 17

[19] Jody C Culham, Stephan A Brandt, Patrick Cavanagh, Nancy G Kanwisher, Anders M Dale, and Roger BH Tootell. Cortical fmri activation produced by attentive tracking of moving targets. *Journal of neurophysiology*, 80(5):2657–2670, 1998. 13, 14

[20] Jeff Donahue, Lisa Anne Hendricks, Sergio Guadarrama, Marcus Rohrbach, Subhashini Venugopalan, Kate Saenko, and Trevor Darrell. Long-term recurrent convolutional networks for visual recognition and description. In *CVPR*, 2015. 3, 13, 14

[21] Jeffrey Donahue, Lisa Anne Hendricks, Sergio Guadarrama, Marcus Rohrbach, Subhashini Venugopalan, Kate Saenko, and Trevor Darrell. Long-term recurrent convolutional networks for visual recognition and description. In *Proceedings of the IEEE conference on computer vision and pattern recognition*, pages 2625–2634, 2015. 3

[22] R Etienne-Cummings and J Van der Spiegel. Neuromorphic vision sensors. *Sensors and Actuators A: Physical*, 56(1-2):19–29, 1996. 3

[23] EV Famiglietti and Helga Kolb. Structural basis for on-and off-center responses in retinal ganglion cells. *Science*, 194(4261):193–195, 1976. 3

[24] Lijie Fan, Wenbing Huang, Chuang Gan, Stefano Ermon, Boqing Gong, and Junzhou Huang. End-to-end learning of motion representation for video understanding. *arXiv preprint arXiv:1804.00413*, 2018. 4

[25] Gunnar Farnebäck. Two-frame motion estimation based on polynomial expansion. *Image analysis*, pages 363–370, 2003. 5, 7, 8

[26] Felix A Gers, Jürgen Schmidhuber, and Fred Cummins. Learning to forget: Continual prediction with lstm. *Neural computation*, 12(10):2451–2471, 2000. 3

[27] James R Golden, Cordelia Erickson-Davis, Nicolas P Cottaris, Nikhil Parthasarathy, Fred Rieke, David H Brainard, Brian A Wandell, and EJ Chichilnisky. Simulation of visual perception and learning with a retinal prosthesis. *bioRxiv*, page 206409, 2017. 3

[28] Kaiming He, Xiangyu Zhang, Shaoqing Ren, and Jian Sun. Deep residual learning for image recognition. In *Proceedings of the IEEE Conference on Computer Vision and Pattern Recognition*, pages 770–778, 2016. 1, 2

[29] Sepp Hochreiter and Jürgen Schmidhuber. Long short-term memory. *Neural computation*, 9(8):1735–1780, 1997. 3

[30] Forrest N Iandola, Song Han, Matthew W Moskewicz, Khalid Ashraf, William J Dally, and Kurt Keutzer. Squeezenet: Alexnet-level accuracy with 50x fewer parameters and< 0.5 mb model size. *arXiv preprint arXiv:1602.07360*, 2016. 2

[31] Xiaolong Jiang, Shan Shen, Cathryn R Cadwell, Philipp Berens, Fabian Sinz, Alexander S Ecker, Saumil Patel, and Andreas S Tolias. Principles of connectivity among morphologically defined cell types in adult neocortex. *Science*, 350(6264):aac9462, 2015. 13, 14

[32] Vadim Kantorov and Ivan Laptev. Efficient feature extraction, encoding and classification for action recognition. In *Proceedings of the IEEE Conference on Computer Vision and Pattern Recognition*, pages 2593–2600, 2014. 3, 7, 8

[33] Hanme Kim, Stefan Leutenegger, and Andrew J Davison. Real-time 3d reconstruction and 6-dof tracking with an event camera. In *European Conference on Computer Vision*, pages 349–364. Springer, 2016. 3

[34] Alex Krizhevsky, Ilya Sutskever, and Geoffrey E Hinton. Imagenet classification with deep convolutional neural networks. In *Advances in neural information processing systems*, pages 1097–1105, 2012. 1, 2

[35] Hildegard Kuehne, Hueihan Jhuang, Estíbaliz Garrote, Tomaso Poggio, and Thomas Serre. Hmdb: a large video database for human motion recognition. In *Computer Vision (ICCV), 2011 IEEE International Conference on*, pages 2556–2563. IEEE, 2011. 4

[36] Yann LeCun, Yoshua Bengio, and Geoffrey Hinton. Deep learning. *Nature*, 521(7553):436–444, 2015. 1

[37] Yann LeCun, Léon Bottou, Yoshua Bengio, and Patrick Haffner. Gradient-based learning applied to document recognition. *Proceedings of the IEEE*, 86(11):2278–2324, 1998. 13, 14

[38] Patrick Lichtsteiner, Christoph Posch, and Tobi Delbruck. A 128x128 120db 15$\mu$s latency asynchronous temporal contrast vision sensor. *IEEE journal of solid-state circuits*, 43(2):566–576, 2008. 3, 5

[39] Min Lin, Qiang Chen, and Shuicheng Yan. Network in network. *arXiv preprint arXiv:1312.4400*, 2013. 13, 14

[40] Jacob Loveless, Sasha Stoikov, and Rolf Waeber. Online algorithms in high-frequency trading. *Communications of the ACM*, 56(10):50–56, 2013. 4

[41] Zhong-Lin Lu and George Sperling. The functional architecture of human visual motion perception. *Vision research*, 35(19):2697–2722, 1995. 14

[42] Niru Maheswaranathan, Stephen A Baccus, and Surya Ganguli. Inferring hidden structure in multilayered neural circuits. *bioRxiv*, page 120956, 2017. 3

[43] Nathan Matsuda, Oliver Cossairt, and Mohit Gupta. Mc3d: Motion contrast 3d scanning. In *Computational Photography (ICCP), 2015 IEEE International Conference on*, pages 1–10. IEEE, 2015. 3

[44] Alex S Mauss, Matthias Meier, Etienne Serbe, and Alexander Borst. Optogenetic and pharmacologic dissection of feedforward inhibition in drosophila motion vision. *Journal of Neuroscience*, 34(6):2254–2263, 2014. 3

[45] Volodymyr Mnih, Adria Puigdomenech Badia, Mehdi Mirza, Alex Graves, Timothy Lillicrap, Tim Harley, David Silver, and Koray Kavukcuoglu. Asynchronous methods for deep reinforcement learning. In *International Conference on Machine Learning*, pages 1928–1937, 2016. 4, 6, 17

[46] Volodymyr Mnih, Koray Kavukcuoglu, David Silver, Alex Graves, Ioannis Antonoglou, Daan Wierstra, and Martin Riedmiller. Playing atari with deep reinforcement learning. *arXiv preprint arXiv:1312.5602*, 2013. 4

[47] Elias Mueggler, Basil Huber, and Davide Scaramuzza. Event-based, 6-dof pose tracking for high-speed maneuvers. In *Intelligent Robots and Systems (IROS 2014), 2014 IEEE/RSJ International Conference on*, pages 2761–2768. IEEE, 2014. 3

[48] R Nelson, EV Famiglietti, and H Kolb. Intracellular staining reveals different levels of stratification for on-and off-center ganglion cells in cat retina. *Journal of Neurophysiology*, 41(2):472–483, 1978. 3

[49] Peter O'Connor and Max Welling. Sigma delta quantized networks. *arXiv preprint arXiv:1611.02024*, 2016. 4

[50] RD Penn and WA Hagins. Kinetics of the photocurrent of retinal rods. *Biophysical Journal*, 12(8):1073–1094, 1972. 4

[51] Javier Sánchez Pérez, Enric Meinhardt-Llopis, and Gabriele Facciolo. Tv-l1 optical flow estimation. *Image Processing On Line*, 2013:137–150, 2013. 1, 7, 8

[52] José Antonio Pérez-Carrasco, Bo Zhao, Carmen Serrano, Begona Acha, Teresa Serrano-Gotarredona, Shouchun Chen, and Bernabé Linares-Barranco. Mapping from frame-driven to frame-free event-driven vision systems by low-rate rate coding and coincidence processing–application to feedforward convnets. *IEEE transactions on pattern analysis and machine intelligence*, 35(11):2706–2719, 2013. 3

[53] Christoph Posch, Daniel Matolin, and Rainer Wohlgenannt. A qvga 143 db dynamic range frame-free pwm image sensor with lossless pixel-level video compression and time-domain cds. *IEEE Journal of Solid-State Circuits*, 46(1):259–275, 2011. 3

[54] Christoph Posch, Teresa Serrano-Gotarredona, Bernabe Linares-Barranco, and Tobi Delbruck. Retinomorphic event-based vision sensors: bioinspired cameras with spiking output. *Proceedings of the IEEE*, 102(10):1470–1484, 2014. 3

[55] Henri Rebecq, Guillermo Gallego, and Davide Scaramuzza. Emvs: Event-based multi-view stereo. In *British Machine Vision Conference (BMVC)*, number EPFL-CONF-221504, 2016. 3

[56] Henri Rebecq, Timo Horstschafer, Guillermo Gallego, and Davide Scaramuzza. Evo: A geometric approach to event-based 6-dof parallel tracking and mapping in real-time. *IEEE Robotics and Automation Letters*, 2016. 3

[57] Christian Reinbacher, Gottfried Munda, and Thomas Pock. Real-time panoramic tracking for event cameras. *arXiv preprint arXiv:1703.05161*, 2017. 3

[58] Sreemananath Sadanand and Jason J Corso. Action bank: A high-level representation of activity in video. In *Computer Vision and Pattern Recognition (CVPR), 2012 IEEE Conference on*, pages 1234–1241. IEEE, 2012. 3

[59] Melissa Saenz, Giedrius T Buracas, and Geoffrey M Boynton. Global effects of feature-based attention in human visual cortex. *Nature neuroscience*, 5(7):631–632, 2002. 14

[60] Christian Schuldt, Ivan Laptev, and Barbara Caputo. Recognizing human actions: a local svm approach. In *Pattern Recognition, 2004. ICPR 2004. Proceedings of the 17th International Conference on*, volume 3, pages 32–36. IEEE, 2004. 13

[61] Karen Simonyan and Andrew Zisserman. Two-stream convolutional networks for action recognition in videos. In *Advances in neural information processing systems*, pages 568–576, 2014. 1, 3, 4

[62] Stelios M Smirnakis, Michael J Berry, David K Warland, William Bialek, and Markus Meister. Adaptation of retinal processing to image contrast and spatial scale. *Nature*, 386(6620):69, 1997. 4

[63] Khurram Soomro, Amir Roshan Zamir, and Mubarak Shah. Ucf101: A dataset of 101 human actions classes from videos in the wild. *arXiv preprint arXiv:1212.0402*, 2012. 4

[64] William K Stell, Andrew T Ishida, and David O Lightfoot. Structural basis for on-and off-center responses in retinal bipolar cells. *Science*, 198(4323):1269–1271, 1977. 3

[65] Gul Varol, Ivan Laptev, and Cordelia Schmid. Long-term temporal convolutions for action recognition. *IEEE Transactions on Pattern Analysis and Machine Intelligence*, 2017. 4, 7, 8, 17

[66] Albert Wang, Sriram Sivaramakrishnan, and Alyosha Molnar. A 180nm cmos image sensor with on-chip optoelectronic image compression. In *Custom Integrated Circuits Conference (CICC), 2012 IEEE*, pages 1–4. IEEE, 2012. 13

[67] Limin Wang, Yuanjun Xiong, Zhe Wang, Yu Qiao, Dahua Lin, Xiaoou Tang, and Luc Van Gool. Temporal segment networks: Towards good practices for deep action recognition. In *European Conference on Computer Vision*, pages 20–36. Springer, 2016. 4

[68] Tianfan Xue, Baian Chen, Jiajun Wu, Donglai Wei, and William T Freeman. Video enhancement with task-oriented flow. *arXiv preprint arXiv:1711.09078*, 2017. 1, 4

[69] Kazumichi Yoshida, Dai Watanabe, Hiroshi Ishikane, Masao Tachibana, Ira Pastan, and Shigetada Nakanishi. A key role of starburst amacrine cells in originating retinal directional selectivity and optokinetic eye movement. *Neuron*, 30(3):771–780, 2001. 3

[70] Joe Yue-Hei Ng, Matthew Hausknecht, Sudheendra Vijayanarasimhan, Oriol Vinyals, Rajat Monga, and George Toderici. Beyond short snippets: Deep networks for video classification. In *Proceedings of the IEEE conference on computer vision and pattern recognition*, pages 4694–4702, 2015. 3

[71] Kareem A Zaghloul and Kwabena Boahen. A silicon retina that reproduces signals in the optic nerve. *Journal of neural engineering*, 3(4):257, 2006. 3

[72] Yi Zhu, Zhenzhong Lan, Shawn Newsam, and Alexander G Hauptmann. Hidden two-stream convolutional networks for action recognition. *arXiv preprint arXiv:1704.00389*, 2017. 4

# Fast Retinomorphic Event-Driven Representations for Video Recognition and Reinforcement Learning - Appendix

## A  Hardware Experiments

Currently, computer vision system for semantic understanding are mostly based on a conventional camera and neural network that are designed for RGB inputs, with a focus on test accuracy, while power consumption and speed are largely under-emphasized. As a result, applications in always-on or embedded vision scenarios are greatly limited. Continuously-on ADC and data transmission regardless of scene context results in wasting 90% total power consumption [66]. As reviewed earlier, the commercially available DVS camera has a similar event-driven mechanism as our proposed EDR. Using the DVS camera could dramatically reduce the energy consumption by up to three orders of magnitude (Table 3), while still capturing the essence of motion dynamics in the scene. (Note that DVS camera outputs do lose some of the features of EDR, such as EMA-based smoothing, soft-thresholding and multiple timescales.) Our hope is that using an event-driven DVS camera and a light-weight deep *recurrent* neural network may yield an energy-efficient imaging system for dynamic event recognition. We set up an early-stage prototype to validate this idea and showcase a hardware based action recognition pipeline.

| Specification | Conventional Intensity Camera (Grasshopper 3) | Event-driven Camera (DVS128) |
| --- | --- | --- |
| Resolution | 2048 x 2048 | 128 x 128 |
| Total Power | 4.5 W | 23 mW |
| Power per video voxel | 11.9 nW / voxel | <0.7 nW / voxel |
| Max. Frame Rate | 90 Hz | 2 kHz (1M events/sec) |
| Dynamic Range | 52.87 dB | 120 dB |
| Max. Bandwidth | 360 Mbps | 4 Mbps |

Table 3: **DVS Hardware Comparison:** Comparison of DVS128 camera and a mainstream intensity camera (adapted from [7]). DVS has significant advantages in energy, speed, dynamic range and data bandwidth.

**Lightweight Network Design**   In order to test EDR in a low power setting, we need to design a *lightweight* network architecture that requires less resources (e.g. depth, width, unrolled timesteps, and GPU memory). Inspired again by biological motion perception, we decide to implement a *feature tracking* ('what target went where' [19]) recurrent convolutional neural (FT-RCN) network for the action recognition task. We base our design on conventional RCNs, but modify the recurrent update to implement the feature tracking functionality. Compared to other existing RCNs, such as Long-term RCNs (LRCN) [20] and Gated Recurrent Unit (GRU) RCNs [5], our proposed EDR-based FT-RCN mimics the recurrent within-channel connectivity observed in visual cortex [31] (Figure 8). In contrast, LRCNs employ all-to-all recurrent connections in the last flattened layer, and so are capable of detecting movement, but have lost the distinction between different features. GRU-RCNs, on the other hand, *do* preserve feature distinctions, but have no recurrent connections between distinct pixels, preventing them from detecting movement. We also explore two standard light-weight ConvNet architectures that provide feature input for the FT-RCN: LeNet [37] and NiN [39]. With EDR as input, these ConvNets extracts motion features and the downstream FT-RCN models individual feature dynamics. Detailed formulation of the FT-RCN are described in Appendix B

**Experiments**   For the experiments, we used the KTH action recognition datasets [60]. We have a random 70/30 training/testing splits on the data. Using the EDR representation layer, events are generated on the fly and passed down to the following network at training. After 100 epoches of training, we show the classification accuracy on the testsets, and show the results below in Table 4.

Once the training is done, we perform transfer learning on the trinaed neural network model to accommodate the DVS hardware input. At testing, we send the DVS input directly to the pretrained ConvNet + RCN without going through the EDR layers. We show the hardware experiment setup in Figure 7a, where we have the DVS128 event camera capture the actions displayed on a computer screen. The action clips are taken from the KTH datasets. We capture 10 event streams for each of the six action classes. The captured event streams are then split by 70/30 ratio. 70% of the event streams are used for fine-tuning the pre-trained model, and the rest 30% are used for testing. For the hardware based testing, We achieve a 72.2% classification accuracy in this six-class classification problem. The confusion matrix of the results are shown in Figure 7b.

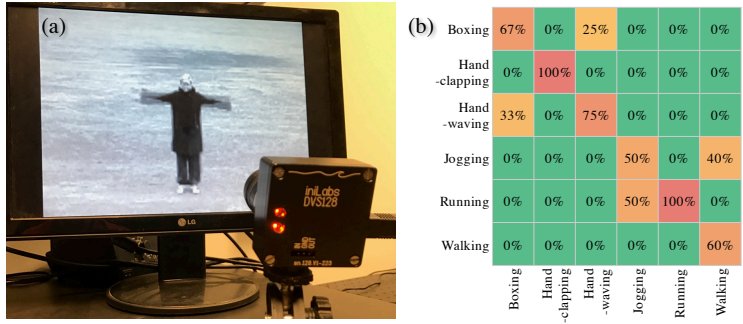

Figure 7: **DVS Hardware Experiment:** (a) Experiment setup.(b) Confusion matrix of the classification results.

## B  KTH Action Recognition Experiments

**Network architectures.**  Our baseline network is a recurrent convolution network (RCN) [20, 5] which takes frame-driven video as input (see Fig. 9), and sends output to a deep ConvNet (DCN). We explore two different types of DCNs, Network-in-Network(NiN)[39] and LeNet[37]. We also explore two different types of late-stage recurrent layers in our RCNs: LSTM-based (LRCN[20]) and GRU-based (GRU-RCN[5]).

**Feature tracking RCN**   Since our proposed EDR extracts temporal features, our consequent recognition pipeline should track such features accordingly. As a matter of fact, a critical function of biological motion perception is feature tracking ("what target went where") [59, 19, 41] Our solution is a feature tracking recurrent convolutional neural (RCN) network to serve as the "visual cortex" for our high-level visual recognition tasks. We based the design on conventional RCNs, but introduce a simple modification to the recurrent update to address the feature tracking functionality. We name of proposed design FT-RCN. We implements this functionality by mimicking aspects of the recurrent within-channel connectivity observed in visual cortex [31].

The diagram of Long-term RCN (LRCN), Gated Recurrent Unit (GRU) RCN and our proposed EDR feature tracking RCN (FT-RCN) are compared against each other in Figure 8. As the figure shows, the differences lie in where the recurrence is introduced, and how recurrence is performed. LRCNs employ all-to-all recurrent connections in the last flattened fully connected layer. They are capable of noticing movement, but have lost the distinction between different features. For GRU-RCN, recurrence happens at each convolutional layer, and is for each same pixel location across all the different feature map. It does preserve feature distinctions, but has no recurrent connections between distinct pixels, preventing it from noticing movement. In conclusion existing LRCN and GRU-RCN architectures have connectivity that makes tracking the movement of stable features difficult/impossible. In contrast, for the proposed FT-RCN, it is similar to GRU-RCN, except that the recurrence is occurring for the entire feature map. This is similar to the like-like connectivity in the visual cortex. Such design will help track high-level EDR features, as the channels in the late stage of DCNs correspond to high-level features.

We build a network that consists of three major components (see Figure 9):

1. *Input Representation Layers:* We choose between our proposed event-driven representation and a conventional frame-driven representation of the input video.

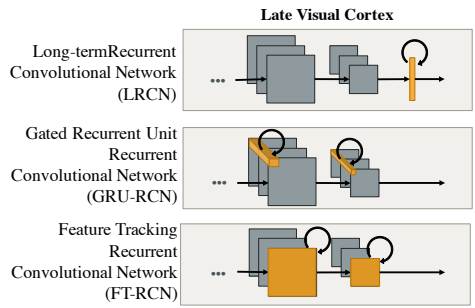

Figure 8: **Comparison of different RCNs:** Late recurrent stages of the proposed FT-RCN , LRCN and GRU-RCN are shown

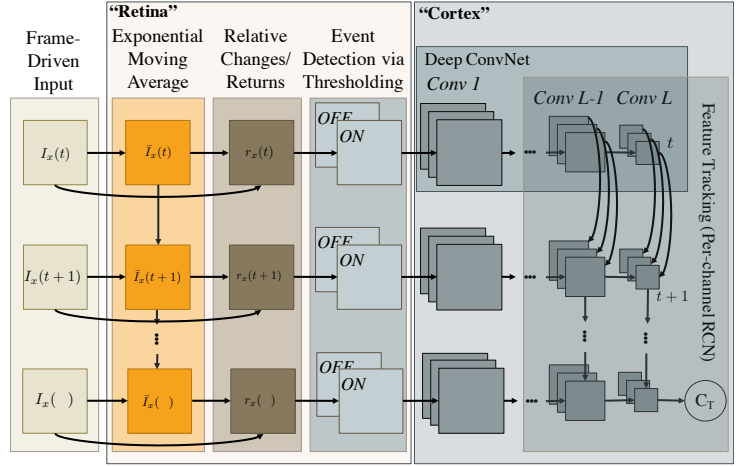

Figure 9: **System Architecture Diagram:** The three major component in our proposed bio-inspired architecture for video recognition. (1) A retinomorphic front-end for generating input events. A neuromorphic back-end cortex that provide high-level semantic understanding of the events, including (2) early convolutional layers and (3) Late recurrent conovolutional layers.

2. *ConvNet Layers:* The ConvNet layers are used for spatial feature extraction. We are able to choose ConvNets of different structures here.
3. *Late-stage Recurrent Layers:* RCNs are used to preserve the temporal context of a dynamic scene. Common RCN architectures includes LRCN and GRU RCN. To this list we also add our proposed feature tracking RCN (FT-RCN) that accepts EDR input.

All together, we have several baseline network architectures that we compare to our EDR-based FT-RCNs. We can assess the value-add of each newly proposed component, and more generally, the value-add of event-driven representations and components.

**Datasets** We choose to focus on action recognition for this paper, but we note in passing that EDR can be applied to *any* video understanding task. We used the KTH datasets, which contains 600 $120 \times 160$ grayscale videos of six action classes, where each class contains 100 videos. Given the videos have various length and due to the GPU memory constraint, we use the first 90 frames of video for experiment. We randomly generate 75/25 train/test splits for use in learning.

**Results and Analysis** Overall, the results are promising: with significant amount of data throughput reduction (i.e. much less activation) and potential computing energy saving (Figure 11), we can observe noticeable improvement in both classification accuracy (Table 4) and training speed (Figure 10), over conventional FDRs, achieving a near-state-of-the-art [1] results at 94.4% accuracy for KTH datasets. Moreover, our proposed FT-RCN "cortex" can nicely handle the event flow, and performs better overall, compared to conventional LRCN. The detailed results and analysis are discussed below.

| RCN Properties | | Input Representation Properties | |
|---|---|---|---|
| Conv Model | Recurrent Structures | Frame-driven Representation | Event-driven Representation |
| LeNet | Long-term RCN | 79.3 % | **86.7 %** |
| LeNet | Feature Tracking RCN | 83.3 % | **88.9 %** |
| NiN | | 90 % | **94.4 %** |

Table 4: **RCN Experiment Results:** Overall, EDR provides better classification results than FDR in RCN on the KTH action recognition datasets. Meanwhile, the proposed FT-RCN seems to be a better cortex for video recognition tasks, as it improves the overall classification performance.

**Input Representation:** *Primitive ON/OFF Event Detector.* In this experiment, we evaluate our primitive event detector, which simulates the ON/OFF pathways between the photoreceptor and the bipolar ganglion cells in the retina. As shown in Table 4, when EDR is introduced as the input representation, we observe that the classification accuracy increases from 79.6%, 83.3% and 90%

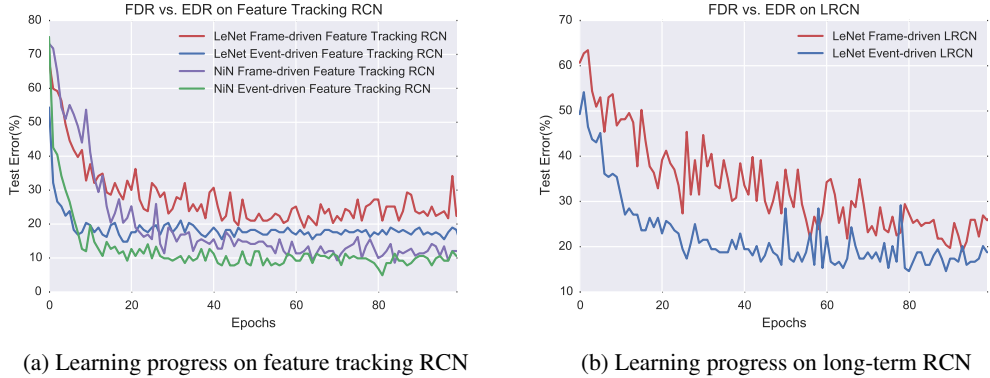

(a) Learning progress on feature tracking RCN

(b) Learning progress on long-term RCN

Figure 10: **Learning progress** of different input representation, ConvNet type and RCN structure. One can observe that EDR results in better classification accuracy and faster converging rate.

(FDR) to 86.7%, 88.9% and 94.4% (EDR), respectively. Moreover, the learning curve shows that learning with EDR is much faster than with the original FDR (Figure 10).

**Recurrent "Cortex" model:** *Feature-Tracking RCN vs. LRCN.* In this experiment, we study the effectiveness of the proposed FT-RCN in handling the EDR feature, compared to a vanilla LRCN. We fixed the input representation layers to EDR and the DCN layer to be LeNet, while setting the late-stage recurrent layers to be FT-RCN and LRCN.

We show the results of different recurrent models in Table 4. EDR-based FT-RCN achieves better performance in the action recognition test (79.3% and 86.7% vs. 86.7% and 88.9%, respectively). We believe the improvement comes from the FT-RCN's ability to better preserving high-level semantic features. As one can see, FT-RCN does a good job accommodating the EDR input, and resolves seemingly more interpretable features compared to the LRCN.

**Convnet Structure:** *NiN vs. LeNet.* In addition, we also explore the effect of different DCN structures. Empirically speaking, deeper DCNs usually provide better end-to-end recognition results, since more layers means more nuisance disentanglement and thus better higher-level feature abstraction. In our experiment, we fix the input to be EDR input, and the RCN to be FT-RCN, and then train end-to-end for the video recognition task using two DCN structures — a shallower LeNet and a deeper NiN network. We show the classification results in Table 4. Unsurprisingly, the NiN-based DCN delivers better overall classification results. Figure 10 shows the learning curve. Interestingly, it seems that EDR provides much more benefit for the shallower LeNet, as compared to the deeper NiN. One possible explanation is that the EDR capture and nuisance more directly, thus the higher-level features becomes more linearly separable. Therefore, even a shallower ConvNet structure will be able to perform more complex tasks well.

**Sparsity, Energy Saving and Computing Efficiency** Compared to the FDR, the EDR provides higher sparsity. We plot the histogram of the activation at different layers of the neural network for both EDR and FDR input in Figure 11. As one can observe, in general, EDR results in a much sparse activation, especially in the early layers. For the activation in the first couple layers in the cortex, EDR results in an tri-mode distribution, which tributes to ON/OFF pathway design, while FDR's activation is further spread out. Furthermore, EDR's sparsity and binary ON/OFF pathway will significantly save the data bandwidth and computing power required.

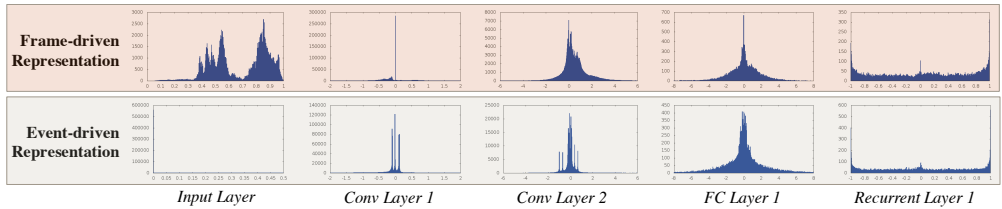

Figure 11: **Activation Distribution in the Artificial Cortex:** Histogram of activation at different layers in EDR and FDR are compared against each other. In general, EDR generates a much sparse activation than FDR. The ON/OFF pathway in EDR produces a visible multimodal distribution in early layers.

## C  RL Experiment Details

**Input Data Dimension**  In the baseline setup, game playing frames are feed into controller network directly. We concatenate EDR with ON and OFF channels and dimension WxHx2 to the original FDR input with dimension WxHx3 in the channel dimension. Thus, new dimension of input is WxHx5 for all our EDR comparison experiments. We refer to this as EDR+FDR for all the Atari based experiments, which is similar to the two-stream framework approach used in video recognition tasks. The difference is that the OF in the conventional two-stream approach is usually pre-computed due to its complexity, here, however, EDR is computed on-the-fly since it's an extremely efficient temporal input representation compared to OF. The intuition behind using two-stream approach in RL is similar to the moviation in the action recognition cases: we would like not only to have a good spatial representation for the context of a given time, but also a temporal representation for the context across time.

**Network Architecture**  After extracting features, we add an additional LSTM layer before feeding the features to the value and policy network branches as we found that this gave better results in our initial analysis. We also replaced LSTM with our feature tracking LSTM (explained in Appendix B) and found that feature tracking LSTM gave better results than using a LSTM in addition to reducing the parameters of the model. So, all the results reported for Atari use feature tracking LSTM.

**RL Training Policies**  For the RL training policy, we experimented with A3C [45] and PAAC [18]. We found that we are able to run more experiments using PAAC because multiple instances of PAAC training can be efficiently paralleled on multiple GPUs. In contrast, in A3C, training agent instances occupy multiple threads with full load, and fill up computation resource quicker than PAAC. The reported experiment results are on PAAC.

In our PAAC experiment setup, we use 32 agents and traverse 5 local steps before aggregating observations into a batch. As a result, each agent in A3C has batch size of 5 and PAAC has batch size of 80. A3C learning rate starts from 0.4226 and decreases exponentially to 0 in 80 million global steps, PAAC learning rate starts from 0.0224 and linear decay to 0 at 80 million global steps. Entropy scaling constant is 0.01 for A3C and 0.02 for PAAC. Both A3C and PAAC training clip gradient based on L2-norm: A3C gradient is clipped at 40.0 and PAAC is clipped at 3.0. All experiments use RMSprop optimizer and discount factor is set to 0.99. Pong training stops at 30 million global steps and training for remaining games stop at 80 million global steps.

## D  UCF-101 Experiment Details

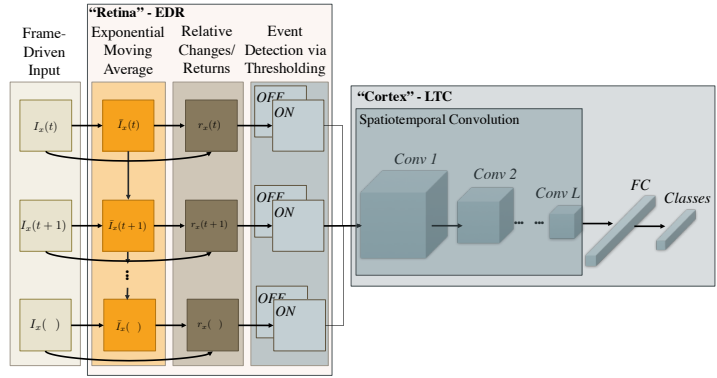

Figure 12: **System Architecture Diagram:** There are two major components in our proposed bio-inspired architecture for video recognition. (1) A retinomorphic front-end for generating input events. (2) A back-end spatiotemporal ConvNet based on the state-of-the art LTC [65] that provide high-level semantic understanding of the events

**Implementation Details**  The end-to-end network architecture is shown in Figure 12. We have two stages of input processing. The first stage happens before training and computes different representations of input videos in their original resolution. The second stage happens during training and involves several data augmentation techniques. First, input videos are scaled into 89x67 pixels.

Second, we randomly sample a volume with size (height, width, 60) from the rescaled input videos, where height and width are randomly chosen from (1, 0.875, 0.75, 0.66) multiples of scaled video size. Randomly sampled volumes are then scaled into network input dimension of 58x58 pixels. Each video is also randomly flipped horizontally with 0.5 chance. Standard evaluation metric is video accuracy. During test time, first 60 frames of test video are used as test clips. Each clip is cropped from its 4 corner and center, forming 5 cropped clips. Each cropped clips is further flipped horizontally and create total 10 cropped clips for each test video. Class result is computed as the maximum of averaged softmax score of all 10 cropped clips. We use stochastic gradient descent as our training algorithm. We treat 9000 video clips as one epoch and stop training after 26 epochs. Learning rate is 1e-3 from epoch 1 to epoch 13, 1e-4 from epoch 14 to 24 and 1e-5 for last two epochs. Batch size is 15.

**EDR Visualizations**  Figure 13a compares a single frame of RGB, EDR and Brox in BreastStroke class. EDR has 60% higher accuracy than Brox and is able to retain the texture of pool line, thanks to the large color gradient between pool lane (red and white) and water (green). Brox on the other hand, failed to extract discernable pattern because moving water creates similar local pixel patches that confuses optical flow algorithm. Figure 13b compares RGB EDR and Brox in HandstandWalking class. Brox is 3x better than EDR and captures only body motions due to overall slow motion speed and stable local patterns. EDR on the other hand, fails short and pickup task irrelevant motion signals due to abundance of color gradient.

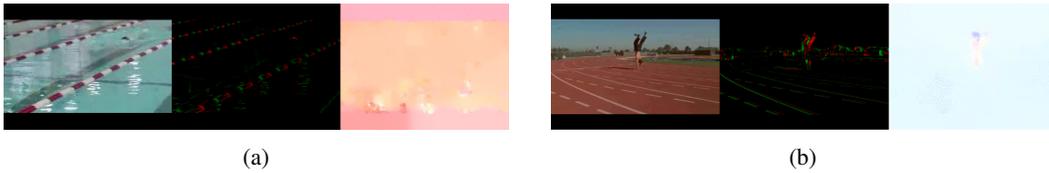

(a)        (b)

Figure 13: **UCF-101 EDR Visualization:** Single frame comparison between original, EDR and Brox [15] for a video in BreastStroke (13a) class and in HandstandWalking (13b) class.